\documentclass[switch]{article}
\usepackage{preprint}
\usepackage{algorithm}
\usepackage{algorithmic}
\usepackage{amsmath, amsthm, amssymb, amsfonts}

\usepackage[numbers,square]{natbib}
\usepackage[utf8]{inputenc}
\usepackage[T1]{fontenc}
\usepackage{xcolor}
\usepackage[colorlinks = true,
            linkcolor = purple,
            urlcolor  = blue,
            citecolor = cyan,
            anchorcolor = black]{hyperref}
\usepackage{booktabs}
\usepackage{nicefrac}
\usepackage{microtype}
\usepackage{lineno}
\usepackage{float}
\usepackage{lipsum}
\usepackage{newfloat}
\DeclareFloatingEnvironment[name={Supplementary Figure}]{suppfigure}
\usepackage{sidecap}
\sidecaptionvpos{figure}{c}

\usepackage{titlesec}
\titlespacing\section{0pt}{12pt plus 3pt minus 3pt}{1pt plus 1pt minus 1pt}
\titlespacing\subsection{0pt}{10pt plus 3pt minus 3pt}{1pt plus 1pt minus 1pt}
\titlespacing\subsubsection{0pt}{8pt plus 3pt minus 3pt}{1pt plus 1pt minus 1pt}

\usepackage{eso-pic}
\usepackage{tikz}
\usepackage{xcolor}
\PassOptionsToPackage{colorlinks=true,linkcolor=gray,urlcolor=gray}{hyperref}

\usepackage{titling}
\usepackage{footmisc}
\usepackage{multirow}
\usepackage{diagbox}
\newcommand{\Author}[2]{\textbf{#1}\textsuperscript{#2}}

\title{SegMix:Shuffle-based Feedback Learning for Semantic Segmentation of Pathology Images}

\author{
  \Author{Zhiling Yan}{1}\and
  \Author{Sicheng Chen}{2}\and
  \Author{Tianyi Zhang}{3} \and
  \Author{Nan Ying}{3} \and
  \Author{Yanli Lei}{3} \and
  \Author{Guanglei Zhang}{3}
}

\date{
  \textsuperscript{1}JD AI Research, Beijing, China \\
  \textsuperscript{2}School of Microelectronics, Xi'an Jiaotong University \\
  \textsuperscript{3}Beijing Advanced Innovation Center for Biomedical Engineering, School of Biological Science and Medical Engineering, Beihang University \\
  [1em]
  \footnotesize \textbf{Corresponding author:} Guanglei Zhang\texttt{<guangleizhang@buaa.edu.cn>} \\
}

\begin{document}
\maketitle
\begin{abstract}
Segmentation is a critical task in computational pathology,
as it identifies areas affected by disease or abnormal growth and is
essential for diagnosis and treatment. However, acquiring high-quality
pixel-level supervised segmentation data requires significant workload
demands from experienced pathologists, limiting the application of deep learning. To overcome this challenge, relaxing the label conditions to image-level classification labels allows for more data to be used and more scenarios to be enabled. One approach is to leverage Class Activation Map (CAM) to generate pseudo pixel-level annotations for semantic segmentation with only image-level labels. However, this method fails to thoroughly explore the essential characteristics of pathology images, thus identifying only small areas that are insufficient for pseudo masking. In this paper, we propose a novel shuffle-based feedback learning method inspired by curriculum learning to generate higher-quality pseudo-semantic segmentation masks. Specifically, we perform patch level shuffle of pathology images, with the model adaptively adjusting the shuffle strategy based on feedback from previous learning. Experimental results demonstrate that our proposed approach outperforms
state-of-the-arts on three different datasets.
\end{abstract}

\vspace{0.35cm}
\section{Introduction}
Accurately locating regions-of-interest (ROIs) is of great importance for pathological diagnosis. Conventionally, pathologists manually screened slides for ROIs and reviewed these regions for tissues with abnormal appearance~\cite{16}, which was time-consuming and exhausting. With the development of deep learning in computer-aided diagnosis, more and more pathologists implement it as a visual aid to highlight and segment regions of diagnostic relevance~\cite{18}. These models mostly rely on high-quality pixel-level annotations, which places much burdens on pathologists. Besides, in clinical diagnosis, we often only have image-level labels instead of pixel-level annotations, since annotated data is often difficult to obtain and is typically not open-sourced.

To save time costs and better exploit the data collected from clinical applications, one approach is to use image-level classification labels to predict pixel-level annotations, known as weakly supervised semantic segmentation~\cite{5,9,20}. ~\cite{11,17} propose sliding patch-based methods to train and predict at the center pixel of a sliding patch to obtain finer predictions. Pixel-based methods ~\cite{1,4,12} typically apply a FCN with contour separation processing to train and predict at the pixel level. HistoSegNet ~\cite{3} is trained on annotated image level of histological tissue type drawn from different organs to segment on pixel level. The prior information, such as thresholds or specific form distributions, is introduced in ~\cite{14,15,25} to improve performance for medical image. However, priors are fairly removed from the diagnosis process of clinical pathology, limiting their applications.

To generate pseudo masks, current methods mostly leverage Class Activation Maps (CAM) ~\cite{24} and the related variants ~\cite{2,19,21,22} to preliminarily locate the target region based on the image-level label, and then iteratively expand and optimize the localization region to achieve semantic segmentation. However, to obtain better segmentation, some characteristics of pathology images need to be further considered, as the previous methods focus on natural images: 
\textbf{(a) Local features.} Factors such as cell size, nucleus-to-cytoplasm ratio, and chromatin distribution, are crucial components of pathology image features. Due to their dispersed nature, careful observation of local areas is necessary. 
\textbf{(b) Global characteristics.} Features including relative cell size, overall cell orientation, and consistent spacing, must be analyzed for the entire image as a whole. 
\textbf{(c) Relative instance relationship.} The relative instance relationships in pathology images include both intra-sample and inter-sample relationships. Intra-sample relationships involve instances of different scales (e.g., cells, cell clusters, and tissues). They are crucial for clinical diagnosis, as pathologists rely on multiscale and multi-angle features rather than single one to make judgments, especially for challenging pathology images. For inter-sample relationships, instances across different samples often have potential connections, since they belong to the same disease. By combining inter-sample instances and allowing information to interact, models can better explore potential relative instance relationships. 
\textbf{(d) Perturbation.} Pathology images are artificially processed, which inevitably introduces noise, leading to the inconsistency in the distribution of pathology images corresponding to the same label. All of these factors can impact on the
generation of semantic segmentation masks.

In this paper, we propose a novel shuffle-based feedback learning method to generate high-quality pseudo masks for weakly supervised semantic segmentation. Specifically, we split the pathology images into patches concerning the granularity of pathology instances and perform the shuffle process across the same batch. Our model self-adaptively changes the learning strategy from coarse to fine by decreasing the patch size and increasing the shuffle ratio of images throughout the learning process. This allows the model to learn multi-scale instance features, including local, global, and relative relationships. The mixed and original images are then used to extract activation information, and the activation region is expanded to generate semantic segmentation masks. In summary, our contribution can be concluded as the following aspects: 
\textbf{(i)} We first
propose a novel shuffle-based feedback learning method to further explore instance relationships in weakly supervised semantic segmentation for pathology images. 
\textbf{(ii)} We devise a feedback learning module inspired by curriculum learning, which makes the model adaptively change the learning strategy with explicit considerations of the multi-scale characteristics of pathology images and the previous learning. 
\textbf{(iii)} Experiments on three datasets illustrate that our algorithm achieves state-of-the-art performance with only image-level annotations.

\begin{figure}[t]
    \centering
    \includegraphics[width=\linewidth]{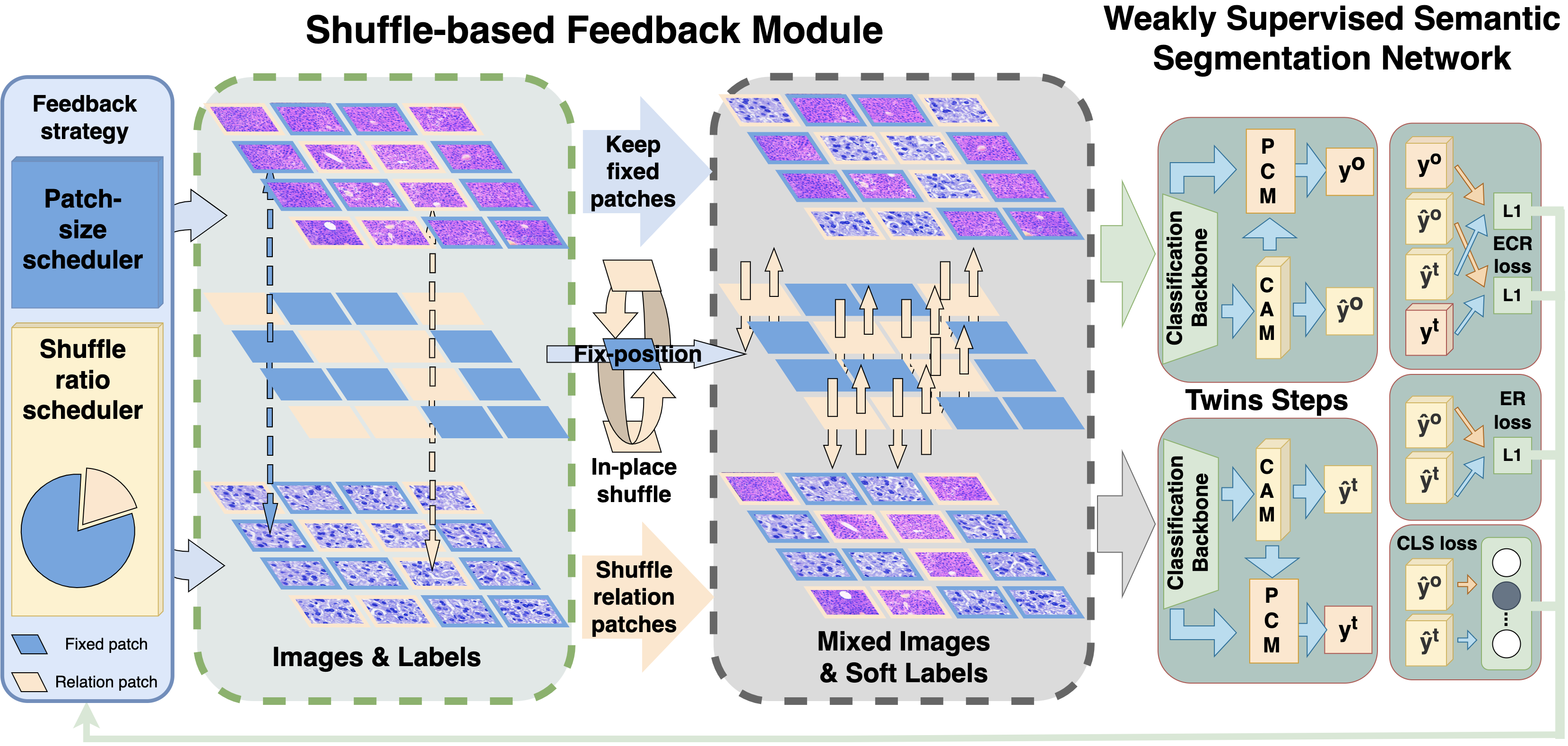}
    \caption{Overview of the proposed method. The model adaptively adjusts the schedulers in the left blue box based on the feedback from previous learning. The patch size scheduler and shuffle ratio scheduler regulate the patch size and shuffle ratio. Then, relation patches are in-place shuffled across the same batch, maintaining their relative positions the same in images. Lastly, the mixed and original images are fed into the Weekly Supervised Segmentation Network for twins’ steps training in the green boxes.}
    \label{fig:main}
\end{figure}
\section{Methods}
In the proposed framework, there are three main components including feedback learning module, shuffle patches and pixel correlation module, as shown in Fig.~\ref{fig:main}. A patch size scheduler and a shuffle ratio scheduler are designed in feedback learning module, as shown in the blue box in Fig.~\ref{fig:main}. With the schedulers, the model adjusts the patch size and shuffle ratio of the images during training based on the learning feedback from previous stages to control the degree of image blending. It enables the model to learn multi-scale pathological features from coarse to fine granularity, therefore better modeling the intra-sample relationships of pathology instances. Then, the model shuffle patches, as shown
in the green and gray boxes in Fig.~\ref{fig:main}. By regrouping patches in in-place shuffle strategy, the method can model the local and global features of pathology images in a better way. The pixel correlation module in the dark green box evaluates the inter-pixel similarities to assist CAM generation.

\subsection{Feedback Learning Module}
\noindent \textbf{Schedulers.} In the framework, we design a patch-size scheduler and a shuffle ratio scheduler. In the patch-size scheduler, we provide a pre-defined list of patch sizes $[p_n, p_{n-1}, \dots, p_1]$, where $p_n > p_{n-1} > \dots > p_1$, and $p_i$ represents the value of the $i^{th}$ patch. Different patch sizes represent different scales of pathology images, and are designed based on the actual size of instances (cells/tissues). For example, when $p_i$ is large, the image is divided according to tissue structures or cell clusters, while the minimum $p_i$ can represent a single cell. For the shuffle ratio scheduler, We provide a growth factor $\alpha$, which is used when the model needs to increase the image shuffle ratio. $f_{i+1} = f_i * \alpha$, where $f_{min} < f_i < f_{max}$, $f_i$ represents the current shuffle ratio, $f_{min}$ and $f_{max}$ denote pre-set thresholds. At the beginning of the training, due to the large patch size and low shuffle ratio, the model learns coarse-grained blended images. Along with the training process, the patch size decreases and the shuffle ratio increases, allowing the model to learn finer features. It is worth noting that we only provide schedulers, and the model itself decides how to select patch size and shuffle ratio based on learning feedback.

\noindent \textbf{Feedback Learning.} Based on the schedulers above, the training model can adjust the mixing scale to control the learning process according to the feedback from previous learning. Specifically, let the loss of the current iteration be $l$, and the learning threshold is $T$. When the loss value of the model is less than a certain threshold, the model has been learned well enough for the current stage, which indicates the feedback from previous learning is positive. Therefore, the model picks a smaller patch size in the scheduler and increases the shuffle ratio with the increasing factor.
\begin{equation}
    p_i \to p_{i+1}; \quad f_{i+1} = f_i * \alpha \quad \iff \quad l < T \quad (1)
\end{equation}
Otherwise, the schedule remains unchanged. And $T$ will decrease along with the training process.

\subsection{Shuffle Patches}
Let $I \in \mathbb{R}^{W \times H \times C}$ represent a pathology image, where $W, H, C$ denote the width, height, and channel of the images respectively. We first break the image into patches according to the patch size $p$ regulated by the patch size scheduler, thus the image can be represented as $I = \{P_0, P_1, \dots, P_n\}$, where $P_i \in \mathbb{R}^{[p, p, C]}$ denotes each patch of the pathology image, and $p$ is the patch size of $P_i$.

Let $f \in [0, 1]$ denote the shuffle ratio from the scheduler. We randomly select $m$ patches in the same position across one batch with $m = [n \times f]$, which are regarded as relation patches $R$, and others remained are fixed patches $F$. Then the relation patches will be in-place shuffled, while the fixed patches hold their positions. The shuffle process is represented as:
\begin{equation}
P_{mix,i} = M \odot P_{F,i} + (1 - M) \odot P_{R,i}, \quad i \in n
\end{equation}
where $\odot$ is element-wise multiplication, $F$ and $R$ are patches from different images. $M$ denotes a binary mask, serving to assign the two kinds of patches. When $i \in R$, $M = 0$, else if $i \in F$, $M = 1$. It is worth noting that, because the shuffle of each patch is independent and completely random, patches in the mixed image may come from multiple images (more than two). The design introduces the relationship perception towards different samples. Meanwhile, We use the same proportion $f$ to blend the labels of these batch samples, which can create soft label to provide a clear image-wise label.

\subsection{Pixel Correlation Module}
We employ a Pixel Correlation Module (PCM) module~\cite{19} to improve the consistency of the prediction. First, we extract an original CAM from backbone model. Then, the cosine distance is calculated to measure the relationship between the adjacent pixels:
\begin{equation}
f(x_i, x_j) = \sigma \left( \frac{\theta(x_i)^T \theta(x_j)}{||\theta(x_i)|| \cdot ||\theta(x_j)||} \right)
\end{equation}
The improved CAM is generated with the weighted sum of the original CAM:
\begin{equation}
Y_p = \frac{1}{\sum \sum (f(x_i, x_j))} \sigma \left( \frac{\theta(X^T)\theta(X)}{||\theta(X)||^2} \right) Y
\end{equation}
where $Y$ is the original CAM, $X$ is the aggregation of some hidden features, $\sigma(\cdot)$ is ReLU function, and $\theta(\cdot)$ is $1 \times 1$ convolution layer.

\section{Experiments}
\subsection{Datasets and Implementation Details}
\noindent \textbf{Datasets.} Three datasets are explored in this paper. The ROSE dataset~\cite{23} contains fast-stained cytopathological images of pancreatic tissues based rapid on-site evaluation (ROSE) technique, including 1,773 pancreatic cancer images and 3315 normal pancreatic cell images. Both the classification labels and segmentation results are confirmed by senior pathologists. The WBC dataset~\cite{10} includes 301 basophils, 1,066 eosinophils, 3,461 lymphocytes, 795 monocytes and 8891 neutrophils. The MARS dataset $^1$ contains 1770 images from the first round of SEED challenge, including 574 normal gastric images, 403 typical tubular adenocarcinoma, and 793 typical mucinous adenocarcinoma.

\noindent \textbf{Implementation Details.} The experiments are fairly done with a same Ubuntu server (Intel(R) Xeon(R) Platinum 8350C CPU and 2 Nvidia RTX3090 GPU). The PyTorch version is 1.10.0, the CUDA version is 11.3 and the python version is 3.8. In each experiment, only one GPU is used. We employed ResNet~\cite{7} as the backbone model for training pseudo segmentation mask generator. Multi-label soft margin loss is adopted for network training. It is trained with optimizer for 8 iterations, with a batch size of 4, an initial learning rate of 0.0001 for ROSE and WBC, 0.00009 for MARS. All datasets are randomly separated into training, validation and test sets following a ratio of 7:1:2. For the evaluation metrics, we adopted the widely used $\text{Dice score} = 2TP/(2TP + FP + FN)$ and $\text{IoU} = target \cap prediction / target \cup prediction$ as in ~\cite{3,6,8,13}. The higher scores means the better performance for both of the metrics.

\begin{table}[htbp]
\centering
\caption{Dice score (\%) and IoU (\%) of methods for weakly supervised semantic segmentation. Ours w/o FL denotes the proposed method without the feedback learning. The best results are in bold.}
\label{tab:comparison}
\begin{tabular}{c|c|c|c|c|c|c|c}
\hline
\multirow{2}{*}{Dataset} & \multirow{2}{*}{Method} & \multicolumn{3}{c|}{Dice Score (DSC) [\%]} & \multicolumn{3}{c}{IoU [\%]} \\
\cline{3-8}
 & & Subtype1 & Subtype2 & Average & Subtype1 & Subtype2 & Average \\
\hline
\multirow{5}{*}{ROSE} & CAM & 39.1 & 52.5 & 45.8 & 32.5 & 38.5 & 35.6 \\
 & SEAM & 35.4 & 69.6 & 52.5 & 28.9 & \textbf{58.3} & 43.7 \\
 & CPN & 42.7 & 62.9 & 50.1 & 33.2 & 51.4 & 42.3 \\
 & Ours w/o FL & 47.6 & 63.3 & 55.4 & 39.9 & 50.8 & 45.4 \\
 & Ours & \textbf{51.4} & \textbf{70.0} & \textbf{60.7} & \textbf{42.4} & 58.1 & \textbf{50.2} \\
\hline
\multirow{5}{*}{WBC} & CAM & 12.1 & 30.7 & 28.3 & 7.0 & 21.3 & 21.0 \\
 & SEAM & 39.5 & 39.0 & 29.8 & 28.9 & 27.2 & 20.4 \\
 & CPN & 45.7 & 36.3 & 31.1 & 33.2 & 24.1 & 22.1 \\
 & Ours w/o FL & 41.0 & 35.2 & 34.6 & 31.6 & 23.7 & 26.6 \\
 & Ours & \textbf{60.9} & \textbf{49.2} & \textbf{41.5} & \textbf{41.4} & \textbf{32.3} & \textbf{29.4} \\
\hline
\multirow{5}{*}{MARS} & CAM & 56.9 & 57.7 & 38.2 & 44.0 & 42.9 & 29.0 \\
 & SEAM & 64.4 & 56.8 & 40.4 & 51.0 & 42.1 & 31.0 \\
 & CPN & 60.2 & 54.1 & 39.8 & 47.5 & 41.8 & 30.8 \\
 & Ours w/o FL & 62.8 & \textbf{58.1} & 40.3 & 49.6 & \textbf{43.7} & 31.2 \\
 & Ours & \textbf{64.5} & 57.2 & \textbf{40.6} & \textbf{51.1} & 42.6 & \textbf{31.3} \\
\hline
\end{tabular}
\end{table}

\subsection{Experimental Results}
\noindent \textbf{Utility Analysis.} The proposed method is compared with other methods including CAM~\cite{24}, SEAM~\cite{19} and CPN~\cite{22}. The result is shown in Tab.~\ref{tab:comparison}. Subtype1 includes the negative samples in ROSE, eosinophils in WBC, and tubular adenocarcinoma in MARS, while Subtype2 denotes positive samples in ROSE, monocytes in WBC, and mucinous adenocarcinoma in MARS. The selected subtypes are the most related to clinical diagnosis. Average is the overall mean value of the three datasets. Tab.~\ref{tab:comparison} shows that our proposed method achieves the best performance, with DSC 60.7\% and IoU 50.2\% on ROSE, DSC 41.5\% and IoU 29.4\% on WBC, DSC 40.6\% and IoU 31.3\% on MARS. In addition, it can be observed that our proposed method performs well on majority of subtypes, particularly on the eosinophil (Subtype1) in WBC, achieving a remarkable improvement of 34.4\% for IoU and 48.8\% for DSC compared to the baseline. To further validate the effectiveness of the Feedback Learning Module, we fix the patch size to 32 and the shuffle ratio to 0.3, and the results are shown in Ours w/o FL. The method without the Feedback Learning Module still outperforms the baselines, which indicates that the shuffle module contributes to segmentation and can better model local instances and global information by regrouping patches. However, with the Feedback Learning Module, the model's performance significantly increases, with up to 6.9\% for DSC and 4.8\% for IoU. This demonstrates the effectiveness of feedback learning, which allows the model to adaptively adjust the shuffle strategy based on the feedback learned previously and better extract features of pathology images. Meanwhile, by continuously adjusting the patch size and shuffle ratio during the shuffle process, the model can learn multi-scale pathological features from coarse to fine granularity, which improves model performance.

\begin{figure}[t]
    \centering
    \includegraphics[width=\linewidth]{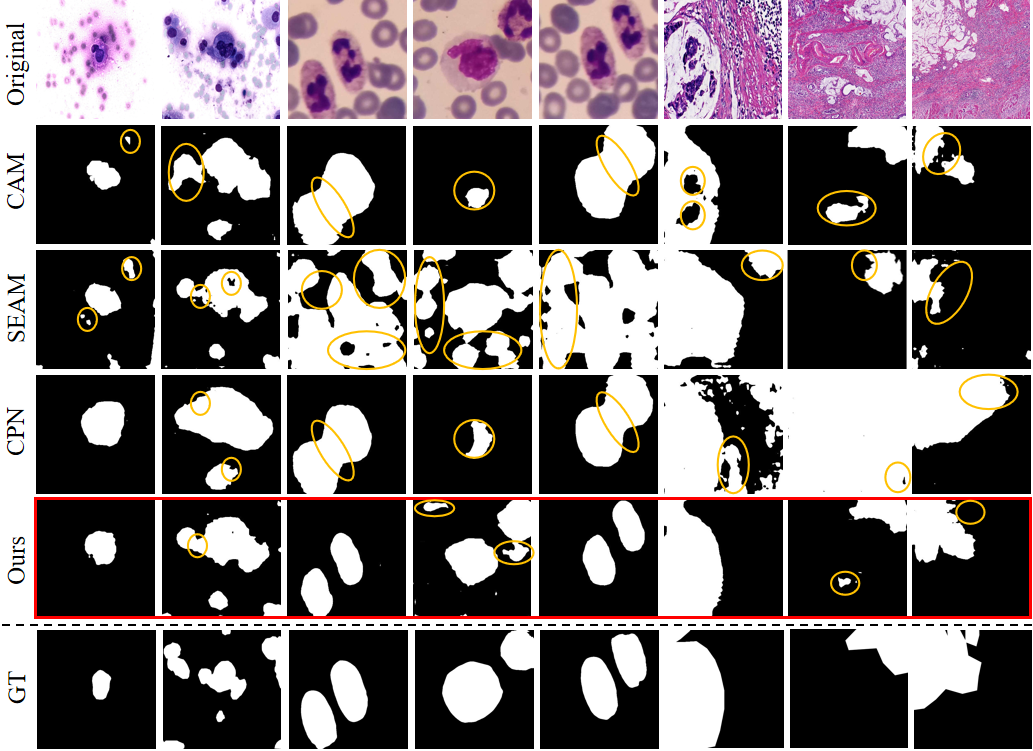}
    \caption{Visualization of crafted examples from different methods. The first row is the original images from ROSE, WBC and MARS. GT represents the ground truth segmentation masks. Other rows are results of weakly supervised methods. The areas with inaccurate predictions are marked with yellow circles.}
    \label{fig:cam}
\end{figure}

\noindent \textbf{Visualization.} We also visualize segmentation results of different methods in Fig.~\ref{fig:cam}. It shows that our proposed method performs the best in most segmentation masks. Baselines tend to highlight and segment regions that contain some irrelevant pixels. It indicates that the previous methods do not have enough capacity to eliminate the interference of perturbation in pathology images, as mentioned in Section 1. By shuffling and regrouping patches, the proposed method reduces the perturbation information in the mixed images, thereby forcing the model to learn the features of the target instance. Based on the utility analysis and visualization result, the proposed method achieves pixel-level annotations of pathology images by leveraging only image-level classification labels.

\begin{table}[t]
\centering
\caption{IoU (\%) of variants with different settings in shuffle and feedback module. Group denotes the shuffled patches are from the same image. Back represents that when the feedback from previous learning is negative, model steps backward to the previous shuffle schedulers. In our proposed method, the shuffled patches are independent, and shuffle schedulers remain the same when feedback is negative.}
\label{tab:ablation}
\begin{tabular}{c|c|c|c|c|c|c|c|c|c}
\hline
\multirow{2}{*}{\textbf{Method}} & \multicolumn{3}{c|}{\textbf{ROSE}} & \multicolumn{3}{c|}{\textbf{WBC}} & \multicolumn{3}{c}{\textbf{MARS}} \\
\cline{2-10}
 & Negative & Positive & Average & Eosino & Mono & Average & Tubular & Mucinous & Average \\
\hline
Baseline & 32.5 & 38.5 & 35.6 & 7.0 & 21.3 & 21.0 & 44.0 & 42.9 & 29.0 \\
Ours-Group & 34.7 & \textbf{59.8} & 47.2 & 39.3 & 30.0 & 28.4 & 51.0 & 42.5 & 31.2 \\
Ours-Back & 39.2 & 58.0 & 48.5 & 40.8 & 32.3 & 29.1 & 51.0 & 42.6 & 31.2 \\
Ours & \textbf{42.4} & 58.1 & \textbf{50.2} & \textbf{41.4} & \textbf{32.3} & \textbf{29.4} & \textbf{51.1} & \textbf{42.6} & \textbf{31.3} \\
\hline
\end{tabular}
\end{table}

\subsection{Ablation Study}
To explore the effects of different settings in shuffle process and feedback module, we conduct the experiments as shown in Tab.~\ref{tab:ablation}. All variants can steadily boost the baseline performance on all benchmarks. In addition to our method with the best performance, other variants lift the baseline result up to 11.6\% and 12.9\% for Group and Back respectively. For positive samples in ROSE, shuffling all patches together performs better than independent mixing strategy, with an improvement of 1.7\%. Although all variants are effective, the overall performance of Group and Back is not the best. Instances across different samples often have potential connections, as they belong to the same disease. However, in the Group strategy, all relation patches from one image are shuffled as a group, which may fail to thoroughly combine inter-sample instances and result in insufficient learning of potential relative instance relationships, negatively affecting performance. As for the Back strategy, since the model needs to return to the previous mixing strategy when the feedback is negative, it is highly likely that the model cannot learn the mixed samples at a fine-grained level. In extreme cases, throughout the training process, only the largest patch size and the smallest shuffle ratio can be implemented to mix images. The decrease in model performance when using the Back strategy indicates that a multi-scale mixing strategy from coarse to fine is necessary to assist the model to learn the features of instances of multi-scales and intra-sample relations. Additionally, poor performance may be attributed to redundant learning and overfitting when stepping backward to the previous shuffle schedulers.
\section{Conclusion}
In this paper, we propose a novel shuffle-based feedback learning method for weakly supervised semantic segmentation on pathology images. By shuffling and regrouping patches containing instances, the model can better mine local and global features of pathology images. Additionally, the model adaptively adjusts the learning strategy through feedback, enabling multi-scale learning of pathological features and instance relationships. On three datasets, our methods significantly outperforms other weakly supervised methods, which achieves pixel-level annotations with only image-level labels. Our next plan is to keep exploring more effective weakly supervised segmentation methods for 3D medical images.

\bibliography{Sections/References}

\end{document}